\documentclass[10pt,twocolumn,letterpaper]{article}

\usepackage[pagenumbers]{cvpr} %

\usepackage{graphicx}
\usepackage{amsmath}
\usepackage{amssymb}
\usepackage{booktabs}
\usepackage[dvipsnames]{xcolor}
\usepackage{soul}
\usepackage{pifont}%
\usepackage{multirow}
\usepackage{xcolor}
\usepackage{enumitem}
\usepackage{tabularx}
\usepackage[accsupp]{axessibility}
\definecolor{green}{RGB}{34, 139, 34}
\definecolor{red}{RGB}{255, 0, 0}
\usepackage[pagebackref,breaklinks,colorlinks]{hyperref}

\usepackage[capitalize]{cleveref}
\crefname{section}{Sec.}{Secs.}
\Crefname{section}{Section}{Sections}
\Crefname{table}{Table}{Tables}
\crefname{table}{Tab.}{Tabs.}

\def\cvprPaperID{49} %
\def\confName{CVPR}
\def\confYear{2023}

\begin{document}

\title{Reliable Student: Addressing Noise in Semi-Supervised 3D Object Detection}

\author{
Farzad Nozarian \and
Shashank Agarwal \and
Farzaneh Rezaeianaran \and
Danish Shahzad \and
Atanas Poibrenski \and
Christian Müller \and
Philipp Slusallek \and
\vspace{-1ex} \\ 
German Research Center for Artificial Intelligence (DFKI) \\
Saarland Informatics Campus \\
{\tt\small \{firstname.lastname\}@dfki.de}
}

\maketitle

\begin{abstract}

   Semi-supervised 3D object detection can benefit from the promising pseudo-labeling technique when labeled data is limited. However, recent approaches have overlooked the impact of noisy pseudo-labels during training, despite efforts to enhance pseudo-label quality through confidence-based filtering.
   In this paper, we examine the impact of noisy pseudo-labels on IoU-based target assignment and propose the \textbf{Reliable Student} framework, which incorporates two complementary approaches to mitigate errors. First, it involves a class-aware target assignment strategy that reduces false negative assignments in difficult classes. Second, it includes a reliability weighting strategy that suppresses false positive assignment errors while also addressing remaining false negatives from the first step. The reliability weights are determined by querying the teacher network for confidence scores of the student-generated proposals.
   Our work surpasses the previous state-of-the-art on KITTI 3D object detection benchmark on point clouds in the semi-supervised setting. On 1\% labeled data, our approach achieves a 6.2\% AP improvement for the pedestrian class, despite having only 37 labeled samples available. The improvements become significant for the 2\% setting, achieving 6.0\% AP and 5.7\% AP improvements for the pedestrian and cyclist classes, respectively. Our code will be released at~\url{https://github.com/fnozarian/ReliableStudent}

\end{abstract}

\section{Introduction}
\label{sec:introduction}
Significant progress has been made in image classification \cite{Imagenet} and object detection \cite{Swin, transformers, Voxelnet, Pointpillars, Second, Pointnet, Pointnetplus, Pvrcnn} with recent developments in deep learning. The availability of large datasets \cite{Imagenet, coco, waymo, once} has helped to accelerate these advancements. However, annotating massive datasets remains a bottleneck, particularly for 2D and 3D object detection. Semi-supervised approaches (SSA) have been proposed to address this problem. Unlike supervised methods, these approaches require only a limited amount of annotated data for training, with the remaining data being unlabeled.

Several semi-supervised techniques have been proposed for object detection, including~\cite{Interactive, unbiased, shanshan, Margingan, humble_teacher, mean_teachers}. Self-training using pseudo-labeling is the most commonly used method and has shown effectiveness in both object detection~\cite{STAC, humble_teacher, unbiased, shanshan} and classification~\cite{Flexmatch, Fixmatch}.
At its core, a student-teacher framework is used to incrementally train teacher and student models on unlabeled data in a mutually beneficial manner. The teacher model is initially trained in a supervised manner on limited labeled data to generate pseudo-labels (PL) to train the student model on unlabeled data. Mean-teacher-based techniques ~\cite{mean_teachers, humble_teacher} use an exponential moving average (EMA) of the student model's weights to update the teacher model's weights, leading to more stable predictions on the unlabeled data.

\begin{figure}
    \centering
    \includegraphics[scale=0.45]{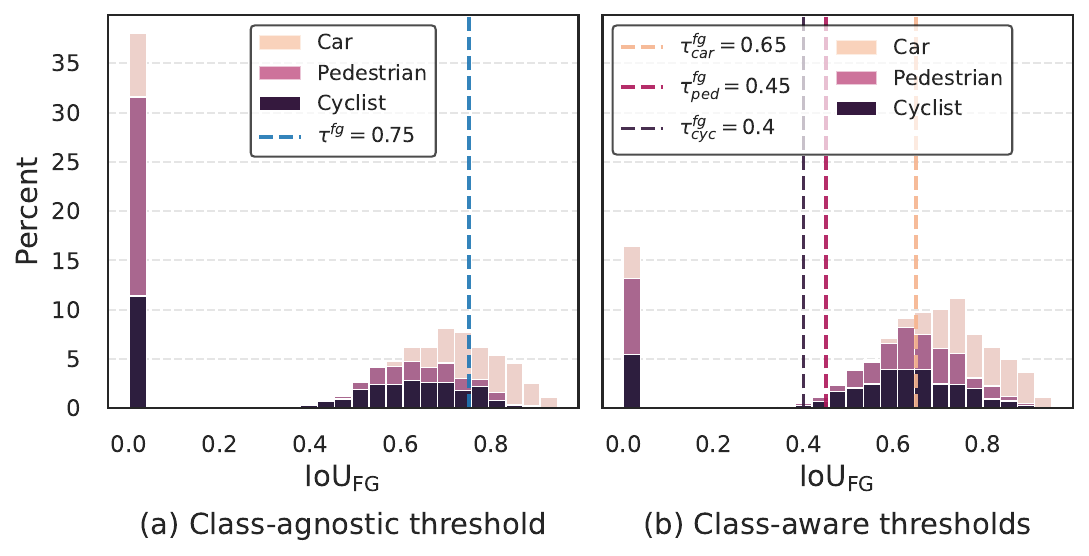}
    \caption{Illustrates the need for class-aware foreground thresholds for foreground/background target assignment. The $\mathrm{IoU_{FG}}$ on the x-axis shows the IoU of proposals with respect to pseudo-labels that are foreground relative to ground truths. (a) The default class-agnostic threshold in the PV-RCNN baseline. (b) Our class-aware thresholds. Lowering the threshold and including more foreground proposals can benefit challenging and uncommon classes.  It also significantly reduces false negatives with IoUs close to zero. (Best viewed in color)}
    \label{fig:teaser}
\end{figure}

Due to its limited pre-training on labeled data, the teacher model fails to generalize effectively, resulting in noisy pseudo-labels that hinder the learning of the student model.
Existing methods overcome this problem by filtering out low-quality pseudo-labels with confidence-based thresholds, acting as a global quality-based filtering mechanism.
However, even with strict filtering, pseudo-labels remain noisy, as shown in Fig.~\ref{fig:teaser} (a).
They have erroneous Intersection over Union (IoU) with proposals that are foreground relative to ground truths.
This poses a significant problem for downstream tasks such as target assignment in Region Proposal Network (RPN) and Region-based Convolutional Neural Network (RCNN) modules, which rely on these noisy IoUs.

The standard target assignment inevitably misclassifies the proposals with IoUs close to zero, \ie, the bar close to the y-axis in Fig.~\ref{fig:teaser} (a), as background, leading to performance degradation.

Fig.~\ref{fig:teaser} also shows distinct class-specific distributions of IoUs due to the different levels of difficulty  and the unbalanced distribution of classes in the dataset.
Neglecting the difference in distributions poses a challenge for class-agnostic target assignment methods in detectors such as PV-RCNN.
A high-value class-agnostic threshold will exacerbate false-negative (FN) errors for difficult classes, such as pedestrians and cyclists, with lower distribution modes, while lowering the threshold will cause many false positives (FP) for the car class, which is easier to learn.

We address these challenges from two perspectives:
1) reducing false-negative and false-positive errors using a new and simple class-aware target assignment approach, and
2) increasing robustness in training against potential failure of our initial assignment by weighting the classification loss to suppress misclassified proposals.
These two steps are complementary, with the first step aiming to minimize assignment errors by considering the difference between the distribution modes of different classes, while the second step mitigates residual errors from the first step.

To this end, we first modify the target assignment process in two key areas where IoU scores are used.
We replace the standard foreground/background random subsampling with a top-k IoU-based subsampler to promote learning from uncertain or difficult background proposals.
We also propose local class-aware foreground thresholds for target assignment. As shown in Fig.~\ref{fig:teaser} (b), the new thresholds include more foreground proposals of difficult classes (leading to higher recall) while preserving a high value for the dominant car class to ensure learning from high-precision proposals.
The foreground and background thresholds divide proposals into three categories: foreground (FG), background (BG), and uncertain (UC).
We assign hard labels to FG and BG proposals and use soft labels for those in the UC category to consider their uncertainty.

Second, to address false negative/positive target assignment errors, we propose to use the teacher to provide reliability scores for the student-generated proposals. To this end, the teacher's RCNN head refines the student's proposals and assigns confidence scores to them, which we use to weight the RCNN classification loss on unlabeled data using different FG/UC/BG weighting options. Our results show that weighting uncertain and background proposals effectively suppresses false positives and false negatives, respectively, and outperforms other proposed weighting schemes.

In summary, our key contributions are as follows:
\begin{itemize}
    \item We thoroughly investigate the impact of noisy pseudo-labels on the IoU-based target assignment.
    \item We propose a class-aware target assignment method to address the target misclassification problem present in recent pseudo-labeling approaches.
    \item We propose different reliability weighting options to suppress false negatives and positives using teacher confidence scores.
    \item We conduct extensive experiments and ablation studies to evaluate the effectiveness of our approach on the KITTI 3D object detection benchmark in a semi-supervised setting.
\end{itemize}

\section{Related Work}
\label{sec:related_work}
\begin{figure*}[h]
\centering
\includegraphics[clip, trim=6.9cm 6cm 6.9cm 6cm, width=0.80\textwidth]{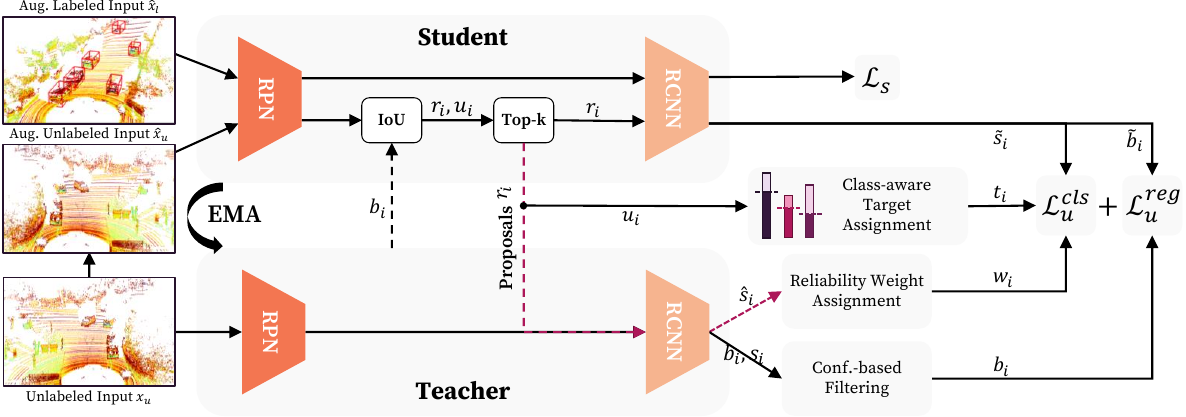}
\caption{\textbf{Overview of our Reliable Student framework.} It uses a teacher-student network, where the EMA teacher produces high-quality pseudo-label boxes $b_i$. We compute the IoU $u_i$ between $b_i$ and the student's post-NMS proposals $r_i$, followed by a top-k sampling of $r_i$ based on $u_i$. The sampled proposals $r_i$ are injected into the student and teacher RCNN heads to predict the objectness scores $\tilde{s}_i$ and $\hat{s}_i$, respectively. 
While $\tilde{s}_i$ serves as an input to the RCNN classification loss $\mathcal{L}^{cls}_u$, $\hat{s}_i$ are converted into reliability weights $w_i$ for $\mathcal{L}^{cls}_u$. The class-aware target assignment module uses thresholds for different classes on $u_i$ to assign objectness targets $t_i$ for $\mathcal{L}^{cls}_u$.}
\label{fig:framework}
\end{figure*}

\subsection{3D Object Detection}
Research on 3D object detection from point clouds focused on a bird's eye view of the lidar point cloud~\cite{multi-view, 3dproposal}. However, VoxelNet~\cite{Voxelnet} employed a different approach by dividing the point cloud into 3D voxels and encoding each voxel using a feature encoding layer. Although 3D convolution layers were applied to further aggregate features, this method was considered time-consuming due to the 3D convolutions involved. To address this, SECOND~\cite{Second} proposed a spatially sparse convolutional network to improve the speed of previous methods. PointPillars~\cite{Pointpillars} then suggested using vertical columns instead of voxels and a 2D convolutional network to encode features. This approach was found to be faster and more robust than previous methods. Another approach by PointNet and PointNet++~\cite{Pointnet, Pointnetplus} was to work directly on encoding points instead of voxels, resulting in more efficient and flexible approaches. In this study, we use PV-RCNN~\cite{Pvrcnn}, a robust two-stage detector that combines the VoxelNet and PointNet approaches and achieves high performance.

\subsection{Semi-Supervised Object Detection}

There have been many studies in the field of semi-supervised 2D object detection.
PseCo~\cite{shanshan} combines both pseudo-labeling and consistency approaches. It uses not only label-level consistency but also feature-level consistency, which further improves the performance of the final detector.
This approach also uses focal loss similar to \cite{unbiased} to alleviate the class imbalance in pseudo-labeling.
\cite{rethinking} considers the localization task as a classification task and proposes a certainty-aware pseudo-label approach.
By quantifying the quality score of classification and regression, they adjust the threshold used for generating pseudo-labels.
Instant-Teaching~\cite{Instant} proposes to generate pseudo annotation for unlabeled data using a weak augmentation in mini-batch, then using these predicted annotations as ground truth of the same image with strong augmentation. For strong augmentation, the authors use Mixup~\cite{mixup}.

Recent works have also focused on class imbalance and confirmation bias issues. LabelMatch~\cite{Chen2022LabelMatching} leverages the labeled data distribution for adaptive thresholding to filter out unbiased pseudo-labels and recalibrates the high-quality unreliable pseudo-labels into reliable ones.
Unbiased Teacher~\cite{unbiased} attempts to address the class-imbalance problem in pseudo-labeling by incorporating a focal loss that forces the model to focus on challenging samples from the underrepresented classes.
Humble Teacher~\cite{humble_teacher} achieves comparable results by using soft labels instead of hard labels with a teacher ensemble network to improve the reliability of the pseudo-labels.

Soft Teacher~\cite{soft} deals with the misclassification of foreground proposals by suppressing the classification loss using the teacher's confidence scores. Our approach follows this but additionally considers the reliability of foreground targets with a foreground reliability weight. Our work also differs from Soft Teacher in that we use a third category of targets in the RCNN, called the Uncertain (UC) region, and assign soft labels to them. These targets may correspond to real foreground or background boxes. Thus, it is crucial to assign appropriate weights to this region to optimize the precision-recall trade-off.
Combating Noise~\cite{wang2021combating} assumes that background proposals are accurate, and it suppresses the noisy foreground proposals losses.
In contrast, we show that dealing with both misclassified foreground and background proposals is important.

There are few works on semi-supervised point-based 3D object detection, such as SESS\cite{zhao2020sess} and 3DIoUMatch\cite{3dioumatch}. SESS uses asymmetric data augmentation techniques and enforces consistency between teacher and student predictions through different losses.
3DIoUMatch~\cite{3dioumatch} proposes a pseudo-labeling approach for both indoor and outdoor 3D object detection.
Inspired by FixMatch~\cite{Fixmatch}, they introduce a joint confidence-based pseudo-label filtering mechanism using predicted objectness and class probabilities.
Additionally, they estimate IoU and use it as a localization quality to filter pseudo-labels.
Unlike 3DIoUMatch, we employ only an objectness threshold, eliminating the complexity of using multiple thresholds.
Moreover, unlike 3DIoUMatch, we adopt objectness supervision on unlabeled data. Our findings indicate that this strategy enhances performance.

\section{Method}
\label{sec:method}

\subsection{Overview}
An overview of our approach is depicted in \cref{fig:framework}. Our approach is based on the mean-teacher framework, where the teacher creates PLs for unlabeled input to serve as a supervised signal for the student.
The student is provided with the strongly augmented version of the unlabeled input as well as the labeled input, and its parameters are updated through backpropagation.
The teacher's parameters, on the other hand, are gradually updated from the student's parameters using the exponential moving average strategy.
To ensure the quality of the generated PLs, we filter them based on their confidence scores.
We introduce the Class-aware Target Assignment module (\cref{sec:contib1}) with class-aware foreground thresholds on IoU of proposals with PLs to improve recall, particularly for challenging classes.
This is based on the understanding that the learning status of classes depends on their difficulty level and the availability of their instances in the dataset.
Given these foreground thresholds and the default background threshold, we define hard classification targets for the foreground and background proposals, while uncertain proposals whose IoUs lay between the FG and BG thresholds are assigned soft targets.

Due to the noisy IoU signal used for target assignment, some proposals may be mistakenly assigned to incorrect targets, leading to FPs and FNs.
To mitigate this, we introduce the reliability-based weight assignment module (\cref{sec:contib2}), which assigns reliability weights to the proposals of each category based on the dominant error type in that category, making the training more robust.
To obtain the reliability weights, we use the teacher model to refine the student's proposals using its RCNN module and use its confidence score $\hat{s}_i$ as additional supervision to improve the student's performance.
Given the student's RCNN refinement box and score $\{ \tilde{b}_i, \tilde{s}_i \}$ and their corresponding targets, we use the teacher score $\hat{s}_i$ to weight the loss of classification on unlabeled data.

\subsection{Class-aware Target Assignment}
\label{sec:contib1}

We investigate the problem of learning from noisy PLs, mainly used to supervise RPN and RCNN modules in the detector.
We focus on the RCNN module and its classification target assignment, where the proposals are assigned with foreground/background labels.

Denote $\mathcal{P} = \{ b_n, c_n, s_n \}^{N_{\mathit{pl}}}_{n=1}$ as the set of filtered PLs consisting of bounding box $b_n$, category label $c_n$, and the confidence score $s_n$.
We define $\{ r_i \}$ as the final proposals or Regions of Interest (RoIs) generated by the student after the IoU-guided filtering and deduplication of RPN proposals using Non-Maximum Suppression (NMS).
Existing pseudo-labeling approaches use the IoU between these RoIs and PLs to assign category labels and FG/BG targets to proposals of unlabeled data in the RPN and RCNN modules of PV-RCNN, respectively.
In RCNN, for a given proposal, if its maximum IoU with PLs, i.e., $u_i = \max_{p \in \mathcal{P}} \mathrm{IoU}(r_i, p)$, exceeds a predefined class agnostic foreground threshold $\tau^{\mathit{fg}}$, it is considered as a foreground proposal.
We define these IoU thresholds used in these two modules as \textit{local thresholds} ($\tau_{\mathit{c}}^{\mathit{fg}}$), as opposed to the \textit{global thresholds} ($\delta_{\mathit{c}}^{\mathit{fg}}$), used to filter out low-quality PLs.

We analyze the suboptimal classification target assignment from PLs with the optimal assignment from GTs. In \cref{fig:teaser}, we evaluate the mean IoU of proposals that are foreground with respect to GTs, i.e., their IoUs with GTs are greater than the evaluation mode class-wise foreground threshold $\Delta^{\mathit{fg}}_{\mathit{c}}$.
We observe two crucial issues when using the standard target assignment.

First, the classes exhibit distinct mean IoU distributions.
Therefore, the standard target assignment strategy based on a single class-agnostic foreground threshold, \eg, $\mathit{\tau^{\mathit{fg}}=0.75}$, cannot reliably classify the proposals.
For the pedestrian and cyclist classes, which have lower distribution modes than the car, such a class-agnostic threshold results in many misclassified foreground proposals whose IoU cannot exceed the threshold by a small margin.
To address this issue, we propose local class-aware foreground thresholds $\tau_{\mathit{c}}^{\mathit{fg}}$, instead of a class agnostic $\tau^{\mathit{fg}}$ on $u_i$ IoUs, to construct the FG/BG target $\mathit{\mathit{t}_{\mathit{i}}}$ for the proposal $\mathit{r}_{\mathit{i}}$ as follows:

\begin{equation}
  t_i  =
    \begin{cases}
      1, & u_i > \tau_{c}^{\mathit{fg}} \\
      \frac{u_i - \tau^{\mathit{bg}}}{\tau_{c}^{\mathit{fg}} - \tau^{\mathit{bg}}} , & \tau^{\mathit{bg}} \le u_i \le \tau_{c}^{\mathit{fg}} \\
      0, & u_i < \tau^{\mathit{bg}}
    \end{cases}.
\label{eqn:cls_target}
\end{equation}

Background proposals have consistently low IoUs, enabling a single class-agnostic threshold $\tau^{\mathit{bg}}$ to distinguish them from other proposals.

Second, the IoUs used for target assignment are unreliable.
This is particularly the case for the pedestrian and cyclist classes, which are difficult to learn due to their object size and the imbalanced class distribution of the dataset. Given the presence of noisy IoUs, despite the implementation of class-specific local thresholds, the assignment carried out in \cref{eqn:cls_target} will inevitably result in the occurrence of false negative (FN) and false positive (FP) errors.

\begin{figure*}[h]
\centering
\includegraphics[scale=0.42]{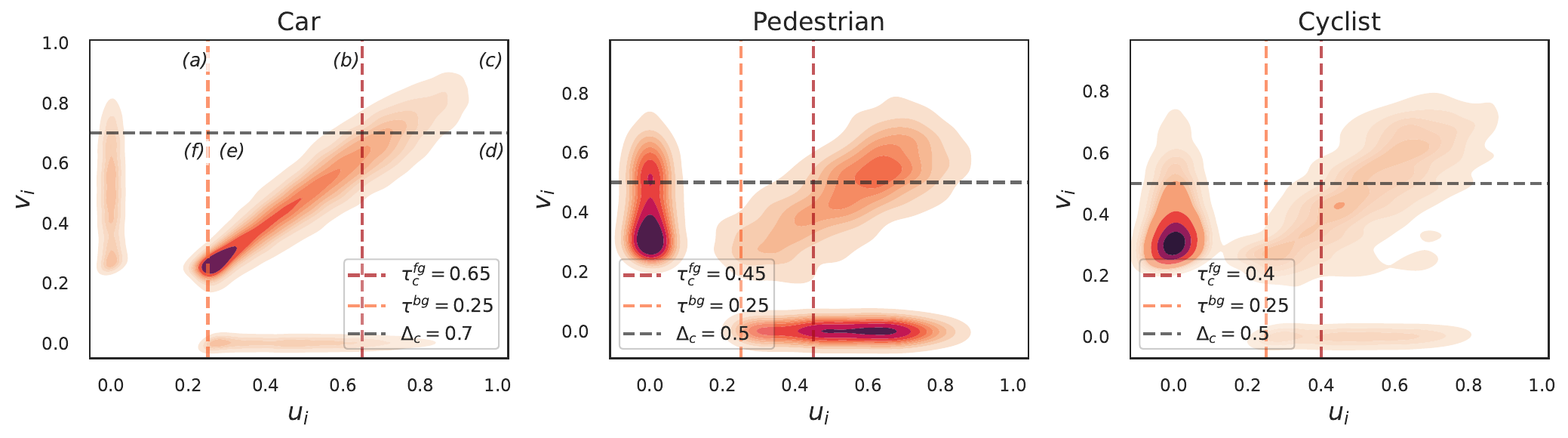}
\caption{Illustrates the density of IoU values of proposals with their matched PL ($\mathrm{u}_\mathrm{i}$) and GT ($\mathrm{v}_\mathrm{i}$) on the x-axis and y-axis, respectively. Denser regions are shown with darker shades. The \textbf{\textcolor{red}{red}} and \textbf{\textcolor{orange}{orange}} vertical lines denote the local foreground (FG) ($\mathrm{\tau^{fg}_{c}}$) and background (BG) ($\mathrm{\tau^{bg}}$) thresholds, while the \textbf{\textcolor{darkgray}{black}} horizontal line represents the FG threshold ($\mathrm{\Delta_{c}}$) for the evaluation mode, dividing the plot into six subregions. Subregions \textit{(a)} and \textit{(f)} represent false negative and true negative proposals, respectively. \textit{(b)} and \textit{(e)} depict proposals lying in the uncertain region and are assigned with soft targets, while \textit{(c)} and \textit{(d)} depict true positive and false positive proposals, respectively. The proposals are obtained from the last few training iterations. We also omit proposals that are in the background with respect to both GT and PL for better visualization. All three plots follow the same subregion breakdown. (Best viewed in color)}
\label{iou_pl_gt}
\end{figure*}

To examine how proposals in the FG, UC, and BG categories are affected by the FP and FN errors, we illustrate the density plots in \cref{iou_pl_gt}, showing the distribution of RoI IoUs relative to both PLs and GTs. The FP proposals are referred to as foreground with respect to PL, but background with respect to GT, whereas those that are the opposite are referred to as FN proposals. As shown, each local class-aware threshold divides the plot into three columns showing FG, UC, and BG sections from right to left.

Ideally, we expect well-calibrated IoU scores such that the IoU of RoIs with respect to PLs are as close as possible to their corresponding IoUs with respect to GTs.
In practice, however, there exist two sub-densities close to the axes contributing to the error.
More specifically, in the foreground region, we observe the density of FP proposals in section (d), near the x-axis, for all classes.
However, for the pedestrian class, we have significantly higher density compared to the other classes.
In the background region, FN proposals are present in (a)  near the y-axis.
The definitions of FP and FN have been extended to the uncertain region, i.e., sections (b) and (e), where FN and FP proposals are located in section (b) and at the bottom of section (e), close to the x-axis, respectively.

\subsection{Reliability-based Weight Assignment}
\label{sec:contib2}
To address these FP and FN erroneous proposals, we focus on making the training robust against a given set of uncertain PLs.
We propose weighting the classification loss of such proposals based on the reliability of their target assignment, i.e., the IoU between RoI and PL.
We seek a reliability score that can consistently assign a low value to both FN and FP proposals.
In this work, we evaluate the reliability score proposed by Soft Teacher. However, any other reliability score can also be plugged into our framework.

We estimate the reliability of the student's proposals based on their corresponding teacher's refined confidence scores. We use these scores to suppress the loss due to FP and FN targets. To this end, we first reverse the augmentation $h$ on the student proposals before sending them to the teacher. The teacher refines each student's proposal $r_i$ using its RoI pooling module and predicts $ \hat{y}_i = \{ \hat{b}_i, \hat{s}_i\}$, where $\hat{b}_i$ and $\hat{s}_i$ denote the corresponding refined bounding box and its confidence score, respectively. The confidence score  $\hat{s}_i$, represents the foreground probability of the refined bounding box proposal, which acts as the reliability score for $r_i$. We propose different reliability weighting schemes based on the teacher's confidence score $\hat{s}_i$, for the RCNN classification loss of unlabeled samples.

Based on our error breakdown in the previous section, we introduce reliability-based weighting options as follows:

\begin{itemize}
    \item \textbf{Background proposals ($\mathrm{BG}$)}: suppress the FN proposals in subregion (f) of \cref{iou_pl_gt} by incorporating the teacher's background score as a weight ($w_i= 1-\hat{s}_i$) for classification loss in subregions (a) and (f).
    
    \item \textbf{Uncertain FN proposals ($\mathrm{UC}_{\mathrm{FN}}$)}: suppress the FN proposals in subregions (b) of \cref{iou_pl_gt} by incorporating the teacher's background score as a weight ($w_i = 1-\hat{s}_i$) for classification loss for subregions (b) and (e).
    
    \item \textbf{Uncertain FP proposals ($\mathrm{UC}_{\mathrm{FP}}$)}: suppress the FP proposals in subregion (e) of \cref{iou_pl_gt} by incorporating the teacher's foreground score as a weight ($w_i = \hat{s}_i$) for classification loss for subregions (b) and (e).
    
    \item \textbf{Foreground proposals ($\mathrm{FG}$)}: suppress the FP proposals in subregion (d) of \cref{iou_pl_gt} by incorporating the teacher's foreground score as a weight ($w_i = \hat{s}_i$) for classification loss for subregions (c) and (d).
\end{itemize}

In all the weighting options, proposals belonging to the remaining categories are assigned with the reliability weight $w_i = 1$. Later in \cref{sec:weighting_types}, we evaluate the application of different weighting options individually and in combination and achieve the best performance from $\mathrm{UC_{FP}+BG}$ by suppressing FPs from uncertain proposals and FNs from background proposals.

We further leverage these reliability-based weights to let the student model learn more about challenging and uncertain proposals instead of the easy backgrounds. The student model's target assignment in RCNN involves computing the IoU between post-NMS proposals and pseudo-labels. Prior works perform sampling on these IoUs such that, at most, 50\% of the foreground proposals are randomly sampled before being passed on for refinement. The remaining background proposals are further randomly subsampled, ensuring that 20\% of them have low IoU ($\eg, <0.1$), that are easily classified as background. Our approach differs in that it avoids subsampling of such easy backgrounds on unlabeled data and instead uses a top-k sampling strategy on the IoU. This allows the model to learn more about the challenging backgrounds.

\begin{table*}[h]
\begin{center}
\scalebox{0.69}{
\begin{tabular}{@{}lcccccccccccccccccc@{}}
\toprule
\multicolumn{1}{c}{\multirow{3}{*}{\textbf{Methods}}} &
  \multicolumn{9}{c}{\textbf{1\%}} &
  \multicolumn{9}{c}{\textbf{2\%}} \\ \cmidrule(l){2-10} \cmidrule(l){11-19} 
\multicolumn{1}{c}{} &
  \multicolumn{3}{c}{\textbf{Car}} &
  \multicolumn{3}{c}{\textbf{Pedestrian}} &
  \multicolumn{3}{c}{\textbf{Cyclist}} &
  \multicolumn{3}{c}{\textbf{Car}} &
  \multicolumn{3}{c}{\textbf{Pedestrian}} &
  \multicolumn{3}{c}{\textbf{Cyclist}} \\ \cmidrule(l){2-10} \cmidrule(l){11-19}  
\multicolumn{1}{c}{} &
  Easy &
  Mod. &
  Hard &
  Easy &
  Mod. &
  Hard &
  Easy &
  Mod. &
  Hard &
  Easy &
  Mod. &
  Hard &
  Easy &
  Mod. &
  Hard &
  Easy &
  Mod. &
  Hard \\ \midrule
PV-RCNN$^\dag$ \cite{Pvrcnn} &
  87.7 &
  73.5 &
  67.7 &
  32.4 &
  28.7 &
  26.2 &
  48.1 &
  28.4 &
  27.1 &
  \textbackslash &
  76.6 &
  \textbackslash &
  \textbackslash &
  40.8 &
  \textbackslash &
  \textbackslash &
  45.5 &
  \textbackslash \\
3DIoUMatch$^\dag$ \cite{3dioumatch} &
  89.0 &
  76.0 &
  70.8 &
  37.0 &
  31.7 &
  29.1 &
  60.4 &
  36.4 &
  34.3 &
  \textbackslash &
  78.7 &
  \textbackslash &
  \textbackslash &
  48.2 &
  \textbackslash &
  \textbackslash &
  56.2 &
  \textbackslash \\ \midrule
PV-RCNN &
  87.6 &
  74.1 &
  67.9 &
  36.5 &
  31.7 &
  28.9 &
  49.9 &
  28.8 &
  27.3 &
  88.9 &
  76.8 &
  71.9 &
  45.1 &
  40.4 &
  35.6 &
  63.0 &
  42.3 &
  38.9 \\ 
3DIoUMatch (Baseline) &
  89.2 &
  76.4 &
  71.3 &
  41.8 &
  35.7 &
  32.9 &
  \textbf{59.9} &
  36.0 &
  33.8 &
  90.7 &
  78.9 &
  74.3 &
  52.9 &
  47.0 &
  41.8 &
  74.2 &
  53.3 &
  49.6 \\
3DIoUMatch + ULB RCNN CLS &
  89.8 &
  76.6 &
  72.0 &
  41.9 &
  36.0 &
  33.1 &
  59.0 &
  35.6 &
  33.3 &
  91.1 &
  79.3 &
  75.3 &
  54.6 &
  48.6 &
  42.8 &
  75.9 &
  54.4 &
  50.7 \\
\textbf{Reliable Student} &
  \textbf{89.7} &
  \textbf{77.0} &
  \textbf{72.5} &
  \textbf{48.0} &
  \textbf{41.9} &
  \textbf{38.4} &
  59.1 &
  \textbf{36.4} &
  \textbf{34.2} &
  \textbf{90.9} &
  \textbf{79.5} &
  \textbf{75.0} &
  \textbf{59.3} &
  \textbf{53.0} &
  \textbf{46.9} &
  \textbf{83.1} &
  \textbf{59.0} &
  \textbf{55.1} \\ \midrule
\% Improvement over Baseline &
  \color{blue}{+0.5} &
  \textcolor{blue}{+0.6} &
  \textcolor{blue}{+1.2} &
  \textcolor{blue}{+6.2} &
  \textcolor{blue}{+6.2} &
  \textcolor{blue}{+5.5} &
  \textcolor{purple}{-0.8} &
  \textcolor{blue}{+0.4} &
  \textcolor{blue}{+0.4} &
  \textcolor{blue}{+0.2} &
  \textcolor{blue}{+0.6} &
  \textcolor{blue}{+0.7} &
  \textcolor{blue}{+6.4} &
  \textcolor{blue}{+6.0} &
  \textcolor{blue}{+5.1} &
  \textcolor{blue}{+8.9} &
  \textcolor{blue}{+5.7} &
  \textcolor{blue}{+5.5} \\ \bottomrule
\end{tabular}}
\end{center}
\caption{Results on the KITTI evaluation set based on mAP over 40 recall positions. PV-RCNN$^\dag$ is the supervised-only baseline, and 3DIoUMatch$^\dag$ is the original work (both based on OpenPCDet v0.3). 3DIoUMatch (Baseline) is our adaptation of the original work to OpenPCDet v0.5, and 3DIoUMatch + ULB RCNN CLS is our modified version of the baseline with objectness supervision from unlabeled data. ($^\dag$) denotes borrowed results from \cite{3dioumatch}, (\textbackslash) indicates non-available results, and \textbf{Bold} indicates the best results from OpenPCDet v0.5.}
\label{tab:baselines}
\end{table*}

Let $\{ \tilde{b}_i, \tilde{s}_i \}$ denote the student's refinement of the proposal $r_i$.
The RCNN classification loss on unlabeled data is summarized as follows:
\begin{equation}
\label{equ::unlabled_cls}
\begin{aligned}
    \mathcal{L}_u^{\mathit{cls}} =
    \frac{\sum_{i}^{N_{b}} w_i l_{\mathit{cls}}(\tilde{s}_i, t_i)}
    {\sum_{i} w_i },
\end{aligned}
\end{equation}
where $N_{b}$ are the total number of proposals for a single unlabeled sample.

Given $N_l$ labeled samples, we define $\mathcal{D}_l = \{(x^l_i, y^l_i)\}^{N_l}_{i=1}$, where $y^l_i$ contains the class labels and bounding box coordinates information, and use $N_u$ unlabeled samples for $\mathcal{D}_u = \{x^u_i\}^{N_u}_{i=1}$. The unsupervised RCNN loss $\mathcal{L}_{u}$ consists of the classification loss $\mathcal{L}_u^{\mathit{cls}}$ from \cref{equ::unlabled_cls}, and box regression loss $\mathcal{L}_u^{\mathit{reg}}$, which is defined as:
\begin{equation}
\mathcal{L}^{\mathit{RCNN}}_u = \frac{1}{N_{u}}\sum^{N_{u}}_{i=1} (\mathcal{L}_u^{\mathit{cls}}(\tilde{s}^u_i, t^u_i) + \mathcal{L}_u^{\mathit{reg}}(\tilde{b}^u_i, b^u_i)), 
\label{eqn:sup_loss}
\end{equation}
where $t^u_i$ is the target for classification loss from \cref{eqn:cls_target}, and $b^u_i$ is the bounding box of the assigned pseudo box based on $u_i$, acting as the regression loss target.
We follow 3DIoUMatch for the RCNN box regression loss $\mathcal{L}_u^{\mathit{reg}}$, as well as for the RPN classification and regression losses, to formulate the unsupervised loss $\mathcal{L}_{u}$.
The supervised loss $\mathcal{L}_{s}$ is calculated similarly on labeled data using ground truth $y^l_i$. The overall loss of the student model is defined as
\begin{equation}
\mathcal{L} = \mathcal{L}_{s} + \lambda_{u}\mathcal{L}_{u}, 
\label{eqn:final_loss}
\end{equation}
where $\lambda_{u}$ is a coefficient balancing the unsupervised loss. The teacher weights are updated as the exponential moving average of the student model.

\section{Experiments}
\label{sec:experiments}

\subsection{Experimental Setup}

We evaluate our method on KITTI \cite{kitti} dataset, consisting of 7,481 training samples and 7,518 test samples. The training samples are divided into the train set (3,712 samples) for training the model and the validation set (3,769 samples) for evaluation.
We use 1\% and 2\% labeled data splits with three folds each, provided by 3DIoUMatch~\cite{3dioumatch}. For each fold, we carry out three trials with different random seed values and report the mean Average Precision (mAP) over all fold-trial combinations.
The mAP is computed using a rotated IoU threshold of 0.7, 0.5, and 0.5 for the car, pedestrian, and cyclist classes, respectively, at 40 recall positions. 
Experiments are conducted over all three object difficulty levels - Easy, Moderate, and Hard.

\subsubsection*{Implementation Details}
For a fair comparison with \cite{3dioumatch}, we utilize PV-RCNN~\cite{Pvrcnn} as the object detection backbone. We used the OpenPCDet v0.5 framework \cite{openpcdet2020} to implement our method and adapted the original 3DIoUMatch from OpenPCDet v0.3 to v0.5 for a fair comparison.
The data augmentation on the student model is based on the 3DIoUMatch settings.
Unlike 3DIoUMatch, which uses both RPN classification and RCNN objectness scores to filter pseudo labels, our approach uses only the RCNN objectness threshold, i.e., $\tau^{\mathit{pl}}_{\mathit{car}}=0.95$ for car, and $\tau^{\mathit{pl}}_{\mathit{ped}}=\tau^{\mathit{pl}}_{\mathit{cycl}}=0.85$ for pedestrian and cyclist.
Unlike 3DIoUMatch, both the RPN and RCNN modules are supervised using labeled and unlabeled data through classification and regression losses, with the unlabeled loss weight $\lambda_{\mathit{u}}=1$. 
On small amounts of data (1\% and 2\%), we pre-train PV-RCNN over 80 epochs with 10 repeated traversals in each epoch and use 60 epochs with 5 repeated traversals in each epoch for the training stage, similar to \cite{3dioumatch}. 
We use a batch size of 8, consisting of 8 labeled and 8 unlabeled samples in both stages. For the evaluation stage, we use the student model.

\begin{figure*}[ht]
\centering
\includegraphics[scale=0.5]{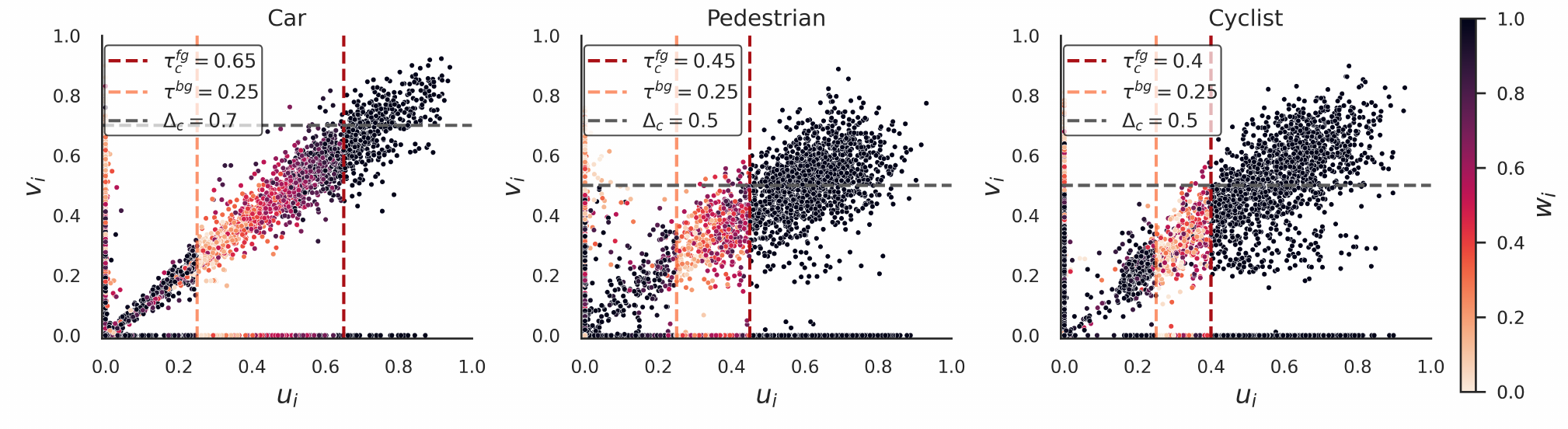}
\caption{Illustrates the assigned reliability weights for RCNN classification loss based on the IoU of the proposals with PLs ($\mathrm{u}_\mathrm{i}$) on the x-axis and GT ($\mathrm{v}_\mathrm{i}$) on the y-axis. The \textbf{\textcolor{red}{red}} and \textbf{\textcolor{orange}{orange}} vertical lines depict the local class-aware foreground (FG) ($\mathrm{\tau^{fg}_{c}}$) and background (BG) ($\mathrm{\tau^{bg}}$) thresholds, respectively, while the \textbf{\textcolor{darkgray}{black}} horizontal line represents the FG threshold ($\mathrm{\Delta_{c}}$) for the evaluation mode. The color bar on the right shows the intensity of the reliability weights. Plots are based on the last few training iterations for better visualization.}
\label{iou_weights}
\end{figure*}

\begin{table*}[h]
\begin{center}
\begin{tabular}{@{}lccccccc@{}}
\toprule
\multicolumn{1}{c}{\multirow{2}{*}{\textbf{Methods}}} &
  \multicolumn{3}{c}{\textbf{1\%}} &
  \multicolumn{3}{c}{\textbf{2\%}} &
  \multicolumn{1}{c}{\multirow{2}{*}{\textbf{\begin{tabular}[c]{@{}c@{}}mAP\\ \% \end{tabular}}}} \\ \cmidrule(lr){2-4} \cmidrule(l){5-7}
\multicolumn{1}{c}{} &
  \textbf{Car} &
  \textbf{Ped.} &
  \textbf{Cycl.} &
  \multicolumn{1}{l}{\textbf{Car}} &
  \textbf{Ped.} &
  \textbf{Cycl.} &
  \multicolumn{1}{c}{} \\ \midrule
Baseline                                            & 76.4          & 35.7          & 36.0          & 78.9          & 47.0          & 53.3          & 54.6 \\ \midrule
$\mathrm{BG}    $                                         & 76.8          & 40.5          & 36.7          &     79.1      &     \textbf{53.2}      &     57.2      &  57.3 (\textcolor{blue}{+2.7})\\
$\mathrm{UC}_{\mathrm{FN}} + \mathrm{BG}$                         & 76.9          & 41.6 & 36.6           &   79.4         &  51.3        &  58.1          &  57.3 (\textcolor{blue}{+2.7})\\
$\mathrm{UC}_{\mathrm{FP}} + \mathrm{BG}$*                                         & \textbf{77.0}          & \textbf{41.9}          & 36.4          &     79.5      &     53.0      &   \textbf{59.0}        &  \textbf{57.8} (\textbf{\textcolor{blue}{+3.2}})\\
$\mathrm{FG} + \mathrm{UC}_{\mathrm{FN}} + \mathrm{BG}$      & 76.8          & 39.9          & \textbf{37.2} & \textbf{79.6} & 53.0          & 55.5          & 57.0 (\textcolor{blue}{+2.4}) \\
$\mathrm{FG} + \mathrm{UC}_{\mathrm{FP}} + \mathrm{BG}$      & 77.0 & 41.4          & 35.9          & 79.5          & 53.2 & 56.8          & 57.3 (\textcolor{blue}{+2.7}) \\ \bottomrule
\end{tabular}
\end{center}
\caption{Ablation study on different reliability-based weighting options on 1\% and 2\% data splits for moderate difficulty level. For a fair comparison, we show the mAP across all classes in the last column, where $\mathrm{UC_{FP}+BG}$ performs the best. (\textbf{*}) indicates our chosen weighting option, and \textbf{Bold} indicates the best results.}
\label{tab:options}
\end{table*}

\subsection{Main Results}
\cref{tab:baselines} shows the results of our approach, the original state-of-the-art 3DIoUMatch method referred to as 3DIoUMatch$^\dag$, and our adapted version of 3DIoUMatch, which is referred to as the baseline. The baseline performs similarly to the original work, except for the cyclist class in the 2\% split, where there is a minor drop of less than 3\%. Note that the baseline does not use the RCNN classification loss on unlabeled data, while our approach benefits from it. Hence, for a more accurate comparison, we have also included the results of our adapted baseline with RCNN classification loss on unlabeled data, which shows an improvement over the naive baseline. We refer to our method as the best option selected from the weighting schemes evaluated in \cref{tab:options},  i.e., $\mathrm{UC_{FP}+BG}$.

Our framework shows superior performance over both 3DIoUMatch and its improved version across all labeled data splits, specially for pedestrian and cyclist classes. While we are also successful in improving for the car class, the margins are relatively small because of two reasons. First, 
the car class suffers from a substantial number of FP errors and in~\Cref{sec:weighting_types}, we show that the effectiveness of reliability weights in such a scenario is limited. Second, the car class being dominant in terms of class distribution is already learnt well in the pre-train stage itself, leaving small room of improvements for the second stage.

\subsection{Ablation Studies}
\subsubsection{Effects of reliability weights}\label{sec:weighting_types}
\begin{figure}
\centering
\includegraphics[scale=0.28]{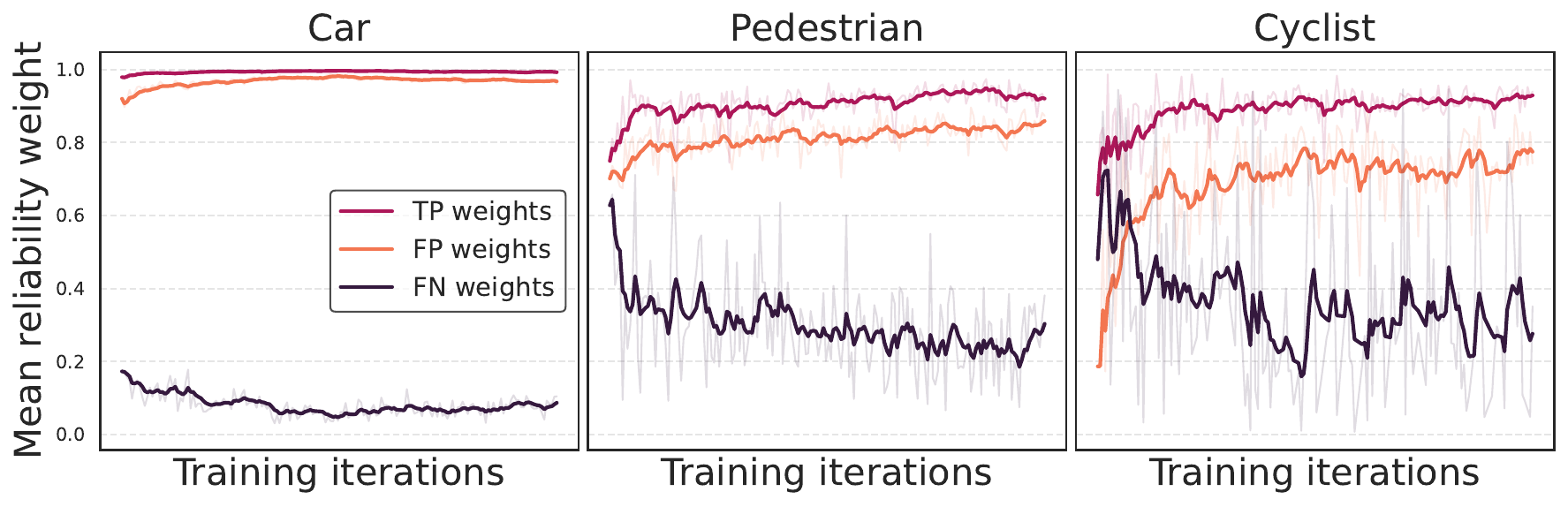}
\caption{Teacher's mean reliability weights, averaged over every few iterations, using the $\mathrm{FG+UC_{FP}+BG}$ weighting type.}
\label{ema_weights_tp_fp_fn}
\end{figure}

\begin{figure}
\centering
\includegraphics[scale=0.35]{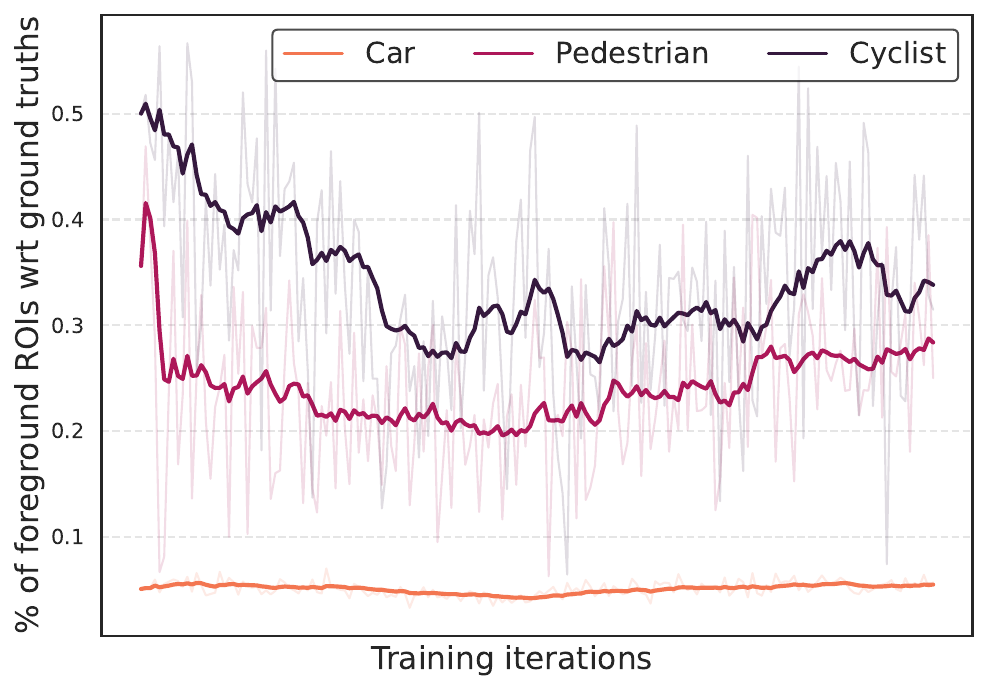}
\caption{Shows the percentage of foreground proposals with respect to GT used to train the FG/BG classification head, highlighting the imbalanced FG/BG ratios across different classes.}
\label{fig::unbalanced_fg_bg_rois}
\end{figure}

\cref{tab:options} ablates the performance over different reliability-based weighting options, improving the mAP over the baseline by 2.7\%-3.2\%.
The $\mathrm{UC}_{\mathrm{FN}}$ and $\mathrm{UC}_{\mathrm{FN}} + \mathrm{BG}$ were evaluated to suppress FN errors, while others assess the effect of suppressing both FN and FP errors. The last two options were assessed to determine efficient ways to weight UC proposals to suppress FN or FP errors. 
While the reliability weights help in all of these options, $\mathrm{UC_{FP}+BG}$ has the highest gain in mAP of 3.2\% over the baseline. Moreover, the teacher's foreground score was found to be more efficient as a weight in the BG option than in the FG option.
We believe that $\mathrm{FG} + \mathrm{UC}_{\mathrm{FN}} + \mathrm{BG}$ has lower performance due to the down-weighting of truly uncertain proposals. 
In \cref{ema_weights_tp_fp_fn}, we show the mean reliability weights of all foreground proposals relative to the PLs with the weighting option of $\mathrm{FG+UC_{FP}+BG}$. As shown, the weights from this option effectively suppress the loss due to FP and FN proposals at the cost of suppressing the loss of some true positives (TP). Moreover, the weights of FPs are relatively higher (close to 1), especially for the car class, and less effective than those for the FNs. We conjecture that this is due to the unbalanced number of FG/BG proposals in the RCNN module. 
\cref{fig::unbalanced_fg_bg_rois} illustrates this by showing the percentage of FG proposals used to train the RCNN classification branch. Note that the car class is highly skewed, with almost 95\% of the proposals as BGs. As a result, the network is biased towards the BG class, and the teacher model cannot provide a reliable FG score for the FP proposals. Whereas, the $\mathrm{UC_{FP}+BG}$ option compensates this by avoiding the suppression of the loss due to the TP proposals, instead mainly suppressing the FPs and FNs, as shown in \cref{iou_weights}.


\subsubsection{Effects of class-aware target assignment}

\cref{tab:class_specific_vs_agnostic} analyzes the effects of local class-aware foreground thresholds over class-agnostic thresholds and their sensitivity to different values.
We show that the class-aware thresholds not only perform better than the default threshold by a large margin, but also they are consistent in performance across different values. We leverage our previous finding that the pedestrian and cyclist classes require lower thresholds than the
car class by adjusting our baseline thresholds by 10\%.

\subsubsection{Effects of top-k based sampler}
\cref{tab:default_sampler_vs_topk} shows that using the balanced random sampler with the class-aware target assignment and unreliability weighting scheme improves the results over the baseline. However, our top-k sampler improves the baseline further by 0.2\%-4.4\% across different classes.

\begin{table}[h]
\centering
\begin{tabular}{@{}llccc@{}}
\toprule
\multicolumn{2}{c}{\textbf{Methods}} & \textbf{Car}  & \textbf{Pedestrian} & \textbf{Cyclist}  \\  \midrule
\multicolumn{2}{c}{Baseline}         & 76.4          & 35.7          & 36.0              \\  \midrule
C-Ag & \multicolumn{1}{c}{0.75}          & 76.6          & 37.0          & 33.2              \\  \midrule
\multirow{3}{*}{C-Aw} & 0.75, 0.55, 0.5                      & 76.5           & \textbf{41.9}          & \textbf{36.6}             \\ 
 &  0.65, 0.45, 0.4*                     & \textbf{77.0}    & \textbf{41.9} & 36.4    \\ 
& 0.55, 0.35, 0.3                     & 76.9           & 41.1          & 36.5             \\ \bottomrule
\end{tabular}
\caption{Ablation study of local class-aware (C-Aw) and class-agnostic (C-Ag) foreground thresholds. C-Aw thresholds are shown for the car, pedestrian, and cyclist (in the same order). We used 1\% labeled data for the moderate difficulty level. (*) indicates our chosen thresholds, and \textbf{Bold} indicates the best results.}
\label{tab:class_specific_vs_agnostic}
\end{table}
\begin{table}[h]
\centering
\begin{tabular}{@{}lccc@{}}
\toprule
\multicolumn{1}{c}{\textbf{Methods}} & \textbf{Car}  & \textbf{Pedestrian} & \textbf{Cyclist}  \\  \midrule
Baseline          & 76.4          & 35.7          & 36.0              \\  \midrule
Default sampler & 76.8          & 37.5                    & 35.5                     \\  %
Top-k sampler           & \textbf{77.0}           & \textbf{41.9}          & \textbf{36.4}             \\  \bottomrule
\end{tabular}
\caption{Ablation study of default random sampler and our top-k sampler. We use 1\% labeled data for the moderate difficulty level.}
\label{tab:default_sampler_vs_topk}
\end{table}

\section{Conclusion}
\label{sec:Conclusion}
Our research on semi-supervised 3D object detection indicates that while generating high-quality pseudo-labels via quality-based filtering is advantageous,
the impact of such noisy pseudo-labels on the IoU-based target assignment module should be considered.
We emphasize the significance of distinct learning curves for different classes and the need for class-specific target assignments, especially with pseudo-labeling techniques.
Moreover, we utilize the teacher model to obtain a reliability score to suppress inaccurate target assignment from noisy pseudo-labels and maintain clear supervision from unlabeled data.
Our research offers an error analysis framework that can be used with other reliability-based metrics to enhance the overall reliability of the system.
We plan to extend it to more autonomous driving datasets and object detectors in the future.

\section*{Acknowledgment}
\label{sec:Acknowledgment}
This work has been funded by the German Ministry for Education and Research (BMB+F) in the project MOMENTUM.

{\small
\bibliographystyle{ieee_fullname}
\bibliography{egbib}
}

\end{document}